\documentclass[conference]{IEEEtran}
\IEEEoverridecommandlockouts
\usepackage{cite}
\usepackage{amsmath,amssymb,amsfonts}

\usepackage{algorithm}
\usepackage{algpseudocode}
\usepackage{graphicx}
\graphicspath{{figures/}}
\usepackage{textcomp}
\usepackage{xcolor}
\usepackage{float}
\def\BibTeX{{\rm B\kern-.05em{\sc i\kern-.025em b}\kern-.08em
    T\kern-.1667em\lower.7ex\hbox{E}\kern-.125emX}}

\DeclareMathOperator{\argmax}{arg\,max} 
\DeclareMathOperator{\maxFun}{max} 

\makeatletter

\def\ps@IEEEtitlepagestyle{%
	\def\@oddfoot{\mycopyrightnotice}%
	\def\@evenfoot{}%
}
\def\mycopyrightnotice{%
	{\footnotesize DOI:10.1109/EST.2019.8806206; 978-1-7281-5546-3/19/\$31.00~\copyright2019 IEEE\hfill}
	\gdef\mycopyrightnotice{}
}

\begin{document}

\title{Hybrid Score- and Rank-level Fusion for Person Identification using Face and ECG Data\\
}

\author{\IEEEauthorblockN{Thomas Truong, Jonathan Graf, Svetlana Yanushkevich}
\IEEEauthorblockA{\textit{Biometric Technologies Laboratory,Department of Electrical and Computer Engineering} \\
\textit{University of Calgary,} Canada \\
\{thomas.truong, jmgraf, syanshk\}@ucalgary.ca}
}

\maketitle

\begin{abstract}
Uni-modal identification systems are vulnerable to errors in sensor data collection and are therefore more likely to misidentify subjects. For instance, relying on data solely from an RGB face camera can cause problems in poorly lit environments or if subjects do not face the camera. Other identification methods such as electrocardiograms (ECG) have issues with improper lead connections to the skin. Errors in identification are minimized through the fusion of information gathered from both of these models. This paper proposes a methodology for combining the identification results of face and ECG data using Part A of the BioVid Heat Pain Database  containing synchronized RGB-video and ECG data on 87 subjects. Using 10-fold cross-validation, face identification was 98.8\% accurate, while the ECG identification was 96.1\% accurate. By using a fusion approach the identification accuracy improved to 99.8\%. Our proposed methodology allows for identification accuracies to be significantly improved by using disparate face and ECG models that have non-overlapping modalities.

\end{abstract}


\section{Introduction}
Biometric technologies have become an important component of many law enforcement and security systems, where efficient and accurate results are of paramount importance \cite{sajjad2017raspberry}. Systems focusing on information privacy and security have similar requirements. In such systems, information from various sources can be fused to form a decision \cite{sajjad2018cnn,wang2008robust}. The basis of this fusion approach is to exploit information available through multiple modalities. Various systems used in the identification process could struggle with identifying different subjects within a database due to sensor errors. However, by focusing on fusing systems that are mutually independent of each other, such as systems that depend on different types of sensors, there should be less overlap of errors in the information each individual systems uses to identify subjects  \cite{sajjad2018cnn}. Therefore, if one systems struggles with an identification of a subject, this should not affect the decision or score another system forms. This means that a system that makes a decision based on multiple sub-systems can be more difficult to fool, making an attack on the identification system harder to execute \cite{faundez2005data}.

In this paper we will present a fusion approach to subject identification using electrocardiogram (ECG) and face modalities. Part A of the publicly available BioVid Heat Pain Database \cite{walter2013biovid,werner2013towards} will be used to design and test these identification models. This part of the dataset consists of synchronized ECG and face data for 87 subjects, each of which has 100 samples. Each sample contains an ECG signal and the synchronized five second video where subjects are exposed to differing levels of heat-pain. The goal of this paper is to introduce a novel hybrid score- and rank- level fusion approach for classification problems involving multiple neural networks. Additionally, to the best of our knowledge, fusion of neural network based ECG and face identification systems on the BioVid dataset has yet to be done. The BioVid dataset is the largest publicly available dataset where both of these modalities are synchronously recorded. The use of both ECG and face identification data to identify individuals should act as a proof-of-concept of how system accuracy can be increased by decentralizing decision scores over multiple disparate sub-systems that have little to no information overlap in terms of the initial sensor measurement.

The paper is outlined as follows: Section \ref{sec:lit} covers the necessary literature review on previous face, ECG, and fusion identification methodologies. Section \ref{sec:method} provides details on our proposed methodology for designing the face, ECG, and fusion identification systems. Section \ref{sec:results} summarizes and discusses the results of our proposed identification system. Section \ref{sec:conclusion} discusses the key outcomes and future works related to our proposed methodology and implementation.

\section{Literature Review} \label{sec:lit}
\subsection{Face Identification Methods}
Methods for identifying faces have been in development for many decades, with initial approaches using concepts such as Hidden Markov Models (HMMs) to encode facial features of different regions of the face \cite{samaria1994parameterisation}. Improvements upon this approach have been made since then, with most modern approaches focusing on convolutional neural networks (CNNs). These networks are composed of layers with different functions, examples of layers can be varying combinations of convolution, pooling, and concatenation layers, with more complex implementations having large networks of these layers. Fundamentally the idea of using CNNs to identify faces is similar to that of HMMs, in that facial features that differentiate individuals are extracted and used to identify individuals.

Collecting enough data to train these networks can be tedious, computationally expensive, and time-consuming. Transfer learning approaches provide a framework to use previously trained networks as a basis for feature extraction, to which additional layers can be added, as needed \cite{pan2010survey}. The feature extraction steps of a pre-trained network used in a transfer learning approach are generally not adjusted, the only aspect of the network that are retrained are the final classification outputs. This approach generally works well as images of any type have similar features that can be extracted such as shapes and color of objects. 

\subsection{ECG Person Identification Methods}
Person identification with ECG signals have been researched extensively in recent years, with many varying (but also all successful) methodologies being developed \cite{Fratini2015}. The ECG signal itself contains many temporal and frequency features which is used to uniquely identify individuals and is covered extensively in \cite{Fratini2015}. For brevity, this paper will omit the technical details on the ECG features. More recently, algorithms using CNNs have been achieving record accuracies for person identifications, reaching up to 100\% identification accuracy on various self-collected and public ECG datasets \cite{Gang2017,Salloum2017}. These proposed CNNs generally involve little pre-processing (in comparison with more traditional methods outlined in \cite{Fratini2015}) and yield excellent results.

\subsection{Fusion Methods for Identification Problems}
Fusion systems, as already briefly mentioned in the introduction, are a popular topic of research in the field of biometrics and identification. Fusion systems involve an assessment of the reliability of the processed information in disparate systems. This assessment is used to create of a fusion algorithm that merges the results of both systems to provide an overall improvement over the individual identification systems \cite{varshney2012distributed}. In theory, is it more difficult to fool multiple identification systems \cite{faundez2005data}, and thus it is of benefit to consider fusion systems for identification system design purposes.

Score-level fusion is the preferable approach for biometric data as classifier scores are easily accessed and processed to be combined \cite{Jain2005}. This is in contrast with sensor or feature-level fusion where the sensor data may not be easily combined (for example, two-dimensional images from a camera and one-dimensional signals from an ECG). Important information which may improve classification results is lost when using decision-level fusion over score-level fusion \cite{Jain2005}. In general, development of score-level fusion methodologies involves determining how much influence each classifier in the fusion system has over the final class output. 

For biometric identification purposes, fusion methodologies have primarily been focused on face, palm, and ear biometrics \cite{Hezil2017,Sarhan2017}. Fusion methods combining ECG and face data have only used more traditional pre-processing and classification methods (such as K-nearest-neighbours and support vector machines) in the past \cite{Israel2004,Boumbarov2012}. State-of-the-art methodologies such as the use of deep neural networks have not been extensively researched for use in a fusion system involving face and ECG data.

This paper aims to propose a novel fusion subject identification algorithm using deep neural networks on ECG and face modalities.

\section{Proposed Methodologies} \label{sec:method}
Our approach involves developing a face identification model (outlined in subsection \ref{subsec:Face}) and an ECG identification model (outlined in subsection \ref{subsec:ECG}). The novelty of our contributions is the fusion algorithm presented in subsection \ref{subsec:Fusion}.

\subsection{Face Identification} \label{subsec:Face}
Several steps were taken to extract and prepare data as input to VGG16 network \cite{simonyan2014very}, the pre-trained network used as a basis in this face identification. First, one image was extracted from each of the 8700 videos. The pre-trained VGG16 network is used to minimize the training time needed for the face identification algorithm. Each image was analyzed via Open CV's \cite{bradski2008learning} Haar like cascade implementation \cite{wilson2006facial} to extract the face present in the image. Due to the nature of the database and subjects being free to move as they wished, sometimes many frames would have to be analyzed before an image with a face present could be extracted.  

Transfer learning was used to classify subjects in Part A of the BioVid Heat Pain Database. The pre-trained VGG16 network \cite{simonyan2014very} was imported with Keras using a Tensorflow backend. A dense layer with 87 outputs, equal to the number of subjects within the database, was added to the output of the VGG16 network. Training on the other layers in the network was turned off to utilize the weights of the original trained network. This minimizes the amount of time needed to train our network.

Images were partitioned into 10 subsets which were used to perform 10-fold cross validation.
 
\begin{figure*}[h!] 
	\centerline{\includegraphics[width=\linewidth]{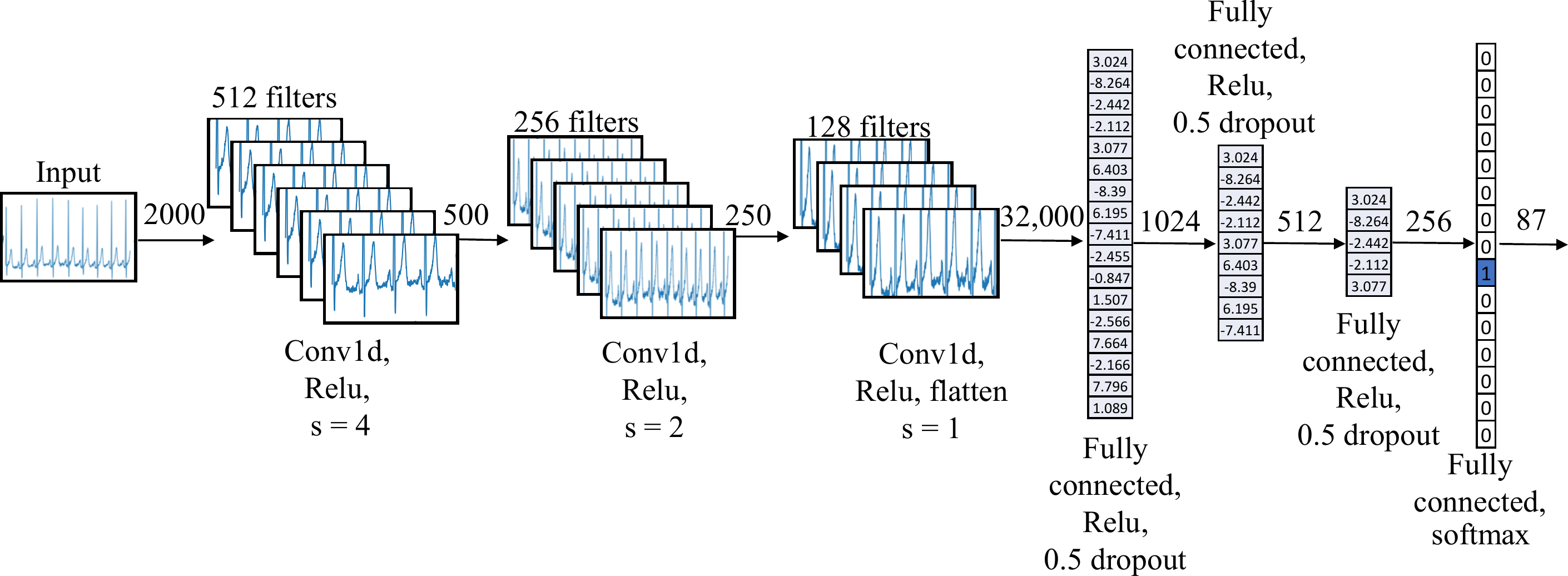}} 
	\caption{Architecture for the ECG person identification network.}
	\label{ecgStruc}
\end{figure*} 

\subsection{ECG Person Identification} \label{subsec:ECG}
The ECG data corresponding to each of the 8700 videos was extracted and processed to have a normalized amplitude with values in [0,1]. Next, the mean of each signal was subtracted so each sample was 0-mean. The first R wave (in general, the R wave peaks are the largest amplitude peaks in an ECG signal) was detected and shifted to $t = 0s$ and the remaining signal was truncated to contain 4 seconds of data. For the BioVid dataset, this means each ECG sample contained 2048 points.

Figure \ref{ecgStruc} shows the architecture for the ECG identification. It is a CNN consisting of 3 convolutional layers and then 3 fully connected layers. The architecture shown was heuristically chosen for maximum accuracy. The fully connected layers are followed by 50\% dropout layers to prevent overfitting during the training. The training and testing uses the same 10-fold cross-validation method used in the face identification system. The training and testing partitions are kept the same as the face identification validation so that the results can be directly compared.


\subsection{Fusion Identification Methodology} \label{subsec:Fusion}

We used 10-fold cross-validation to have 10 models for each system. The partitioning of the data during the 10-fold cross-validation was kept the same for both systems so the models can be paired up. For example, the first model for the face identification system was trained and tested on the same data as the first model of the ECG identification system, as such the results can be directly combined and compared.

\begin{figure}[h!]
	\centering
	\hspace{6mm}
	\includegraphics[width=\linewidth]{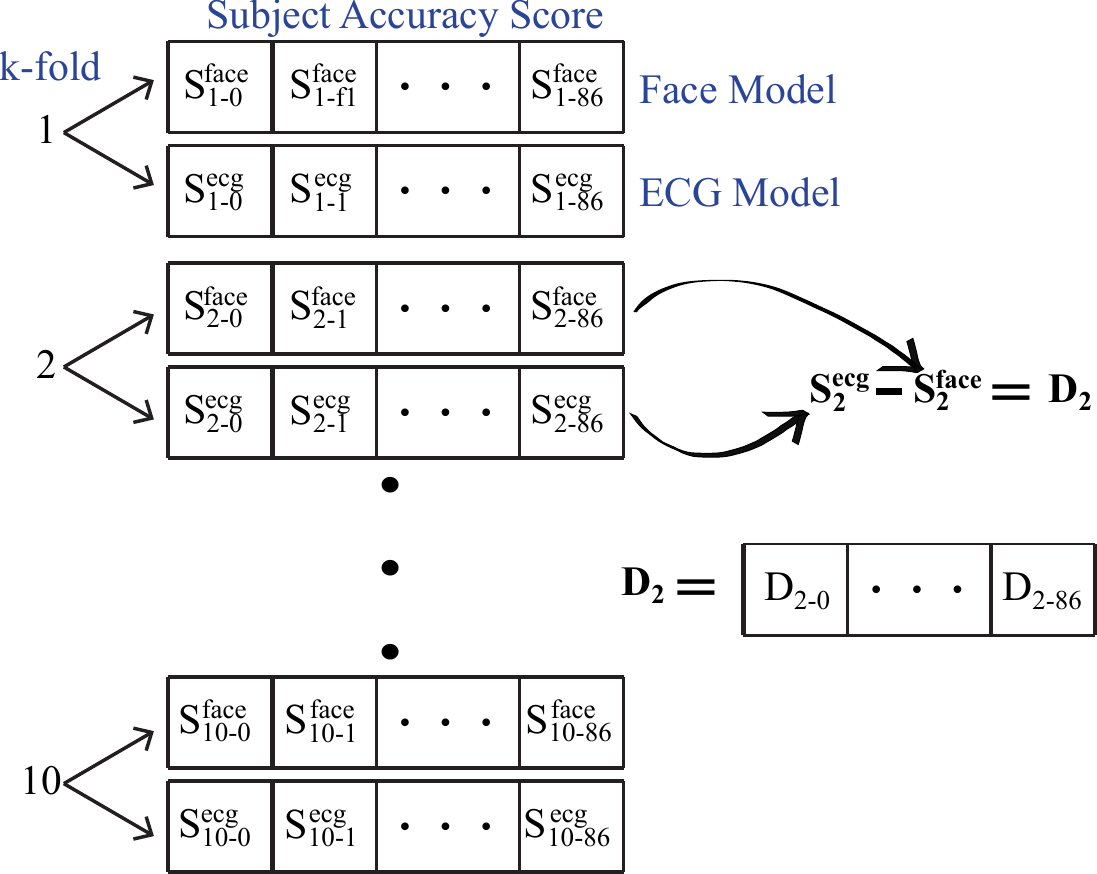} 
	\caption{The results of 10-fold cross validation for each model. Each fold produces a \textit{subject accuracy score} $\textbf{S}$ for each model.}
	\label{fig:subjectAccuracy}
\end{figure}

We use a hybrid score- and rank-level fusion algorithm for our fusion identification system. The fusion network uses two types of scores for each model to compute a final decision.

\begin{figure*}[h!]
	\includegraphics[width=\textwidth]{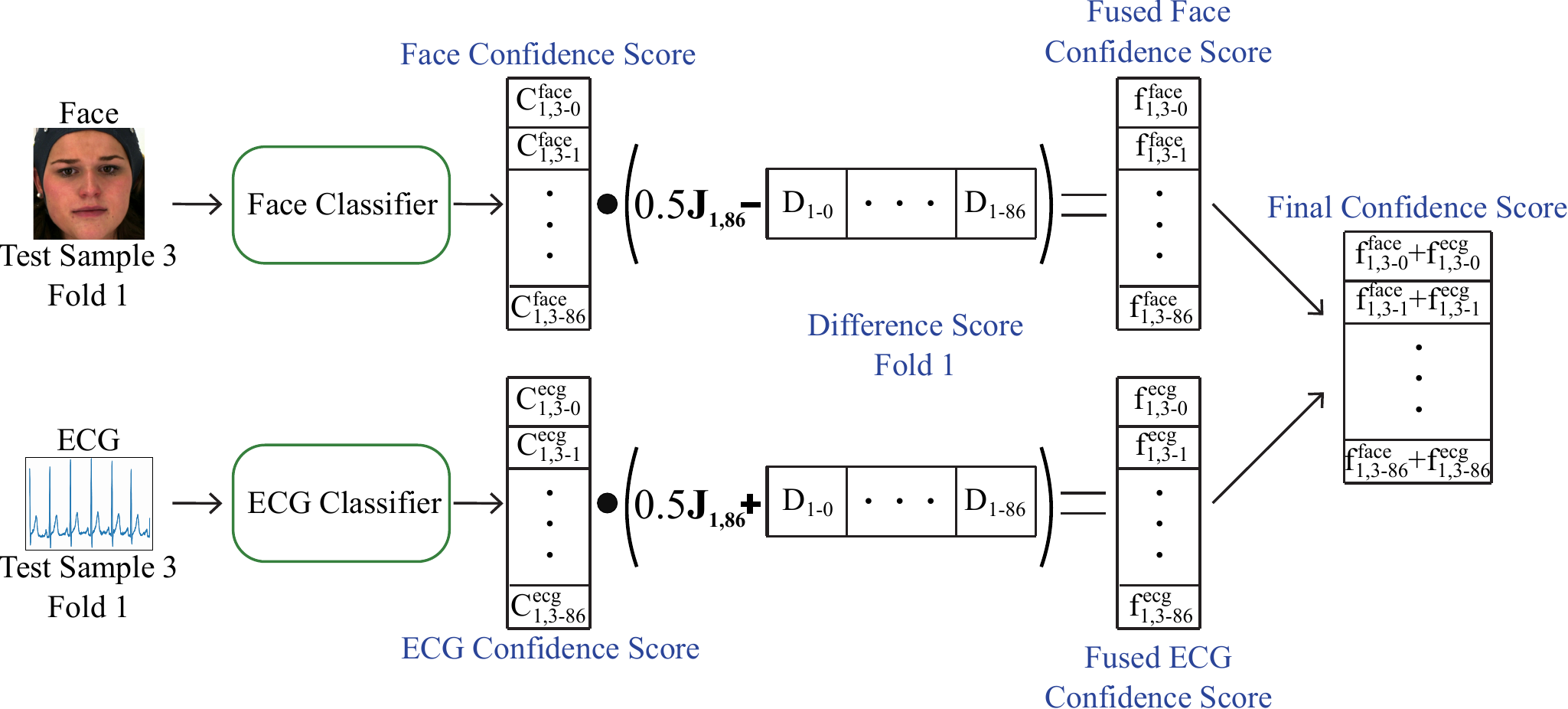} 
	\caption{Diagram showing the process flow of calculating a \textit{final confidence score}, $\textbf{F}$, vector. The classified subject is given by taking $\argmax(\textbf{F})$ and is found to be subject 5.}
	\label{fig:fusionAlgorithm}
\end{figure*}

The first score is the \textit{model confidence score}. Each model has an 87 length vector, $\textbf{C}$, output containing values which can be interpreted as a vector of confidence values. Each sample that is fed into the identification model has its own vector of confidence values. $\textbf{C}^{\text{ecg}}_{k,n}$ and $\textbf{C}^{\text{face}}_{k,n}$ refer to the $\textbf{C}$ vectors for the $k^{th}$ fold and $n^{th}$ sample for the respective ECG and face models. $C^{\text{face}}_{k,n-i}$ and $C^{\text{ecg}}_{k,n-i}$ are the values in the $i^{th}$ element in the for $k^{th}$ fold and $n^{th}$ sample in $\textbf{C}$. There are 8700 different vectors $\textbf{C}$ for each model since the BioVid dataset contains 8700 samples. Since there are 20 models for each fold (1 for face and 1 for ECG, per fold) and 8700 samples per model, we have $20 \times 8700 = 174,000$ distinct $\textbf{C}$ vectors. During the training of the face and ECG models, $\textbf{C}$ was fed through a softmax activation layer for classification. For the fusion system, we remove the softmax activation layer to use the values of $\textbf{C}$. Let  $\argmax_n(\textbf{A})$ be the index for the $n^{th}$ largest element in any one-dimensional vector $\textbf{A}$. $\argmax_1(\textbf{C})$ provides the first ranked subject prediction for the respective model and sample.  $\argmax_2(\textbf{C})$ through $\argmax_5(\textbf{C})$ provides the second through fifth ranked subject predictions. Similarly, let $\maxFun_n(\textbf{A})$ be the value of the $n^{th}$ largest element in any one-dimensional vector $\textbf{A}$.

The second score is the \textit{subject accuracy score}, $\textbf{S}$. $\textbf{S}$ is another 87 length vector. Each element in $\textbf{S}$ contains the score for the subject with subject number corresponding to the element index. For each model we can calculate a score on a per-subject basis and there are 87 subjects with 90 training samples per subject for this dataset. $\textbf{S}^{\text{ecg}}_{k}$ and $\textbf{S}^{\text{face}}_{k}$ refer to $\textbf{S}$ for the $k^{th}$ fold for their respective ECG and face models . $S^{\text{face}}_{k-i}$ and $S^{\text{ecg}}_{k-i}$ are the values in the $i^{th}$ element in the for $k^{th}$ fold $\textbf{S}$ vector. A visual representation of the structure of the vectors is depicted in Figure \ref{fig:subjectAccuracy}. The algorithm used to calculate $\textbf{S}^{\text{ecg}}_{k}$ is shown with Algorithm \ref{alg:scoreCalc}.

\begin{algorithm}
	\caption{Score calculation algorithm for fold k of the ECG model, $\textbf{S}^{\text{ecg}}_{k}$}\label{alg:scoreCalc}
	\begin{algorithmic}[1]
		\State $\textbf{Y} \gets \text{true subject numbers for fold k training set}$
		\State  $\textbf{S}^{\text{ecg}}_{k} \gets 0\cdot\textbf{J}_{1,M}$ \Comment{M is number of classes}
		\For{$n = 0$ to $N$}\Comment{N samples in training set}
		\State $\textbf{C}^{\text{ecg}}_{k,n} \gets \textit{model confidence scores} \text{ for sample }n$
    	\If{$\argmax_1(\textbf{C}^{\text{ecg}}_{k,n})=Y_n$} 
			\State $\text{confDiff} \gets 0$ 
		\ElsIf {$\argmax_2(\textbf{C}^{\text{ecg}}_{k,n})=Y_n$} 
			\State $\text{confDiff} \gets \maxFun_1(\textbf{C}^{\text{ecg}}_{k,n}) - \maxFun_2(\textbf{C}^{\text{ecg}}_{k,n})$
		\ElsIf {$\argmax_3(\textbf{C}^{\text{ecg}}_{k,n})=Y_n$} 
			\State $\text{confDiff} \gets \maxFun_1(\textbf{C}^{\text{ecg}}_{k,n}) - \maxFun_3(\textbf{C}^{\text{ecg}}_{k,n})$
		\ElsIf {$\argmax_4(\textbf{C}^{\text{ecg}}_{k,n})=Y_n$} 
			\State $\text{confDiff} \gets \maxFun_1(\textbf{C}^{\text{ecg}}_{k,n}) - \maxFun_4(\textbf{C}^{\text{ecg}}_{k,n})$
		\ElsIf {$\argmax_5(\textbf{C}^{\text{ecg}}_{k,n})=Y_n$} 
			\State $\text{confDiff} \gets \maxFun_1(\textbf{C}^{\text{ecg}}_{k,n}) - 	\maxFun_5(\textbf{C}^{\text{ecg}}_{k,n})$
		\Else 
			\State $\text{confDiff} \gets 1.0$
		\EndIf
		\State $S^{\text{ecg}}_{k-Y_n} \gets S^{\text{ecg}}_{k-Y_n}  + 1.0 - \text{confDiff}$
		\State $S^{\text{ecg}}_{k-\argmax{(\textbf{C}^{\text{ecg}}_{k,n}})} \gets S^{\text{ecg}}_{k-\argmax{(\textbf{C}^{\text{ecg}}_{k,n}})} - \text{confDiff}$
		\If{$S^{\text{ecg}}_{k-\argmax{(\textbf{C}^{\text{ecg}}_{k,n}})} < 0$} 
			\State $S^{\text{ecg}}_{k-\argmax{(\textbf{C}^{\text{ecg}}_{k,n}})} \gets 0$ 
		\EndIf
		\EndFor
		\State  $\textbf{S}^{\text{ecg}}_{k} \gets \textbf{S}^{\text{ecg}}_{k}\big/\big(N/M\big)$
	\end{algorithmic}
\end{algorithm}

Algorithm \ref{alg:scoreCalc} checks the top five ranked predictions from the ECG model and calculates a confidence score difference, \textit{confDiff}. \textit{confDiff} is the difference between the most confident prediction and the correct prediction. The larger \textit{confDiff} is, the less confident the model is on the true (correct) subject's class. Line 18 of Algorithm \ref{alg:scoreCalc} awards subjects that are correctly identified in the top 5 ranked predictions. If the true subject is not present within the top five ranked predictions of the model, \textit{confDiff} is assigned a value of 1.0 and no score is awarded to the model. Additionally, Line 19 of Algorithm \ref{alg:scoreCalc} reduces the score any subjects falsely identified by the model using \textit{confDiff} to punish false identifications. Note that $\textbf{J}_{nn,mm}$ is the matrix of ones with dimensions nn$\times$mm. Algorithm \ref{alg:scoreCalc} is used to calculate $\textbf{S}^{\text{face}}_{k}$ as well, replacing $\textbf{S}^{\text{ecg}}_{k}$ with $\textbf{S}^{\text{face}}_{k}$ and $\textbf{C}^{\text{ecg}}_{k,n}$ with $\textbf{C}^{\text{face}}_{k,n}$.

We calculate a difference between these scores, $\textbf{D}_{k} = \textbf{S}^{\text{ecg}}_{k} - \textbf{S}^{\text{face}}_{k}$. $\textbf{D}_{k}$ contains values which describe how well each model performs relative to the other model. A positive value for $D_{k-i}$ indicates that the ECG model is, loosely speaking, "better" than the face model at correctly identifying subject $i$. Similarly, a negative value for $D_{k-i}$ indicates that the face model is better than the ECG model at correctly identifying subject $i$. $\textbf{D}_{k}$ can be normalized if desired. We empirically determined that normalizing $\textbf{D}_{k}$ to be within $[-0.20,0.20]$ provided the best results in our experiment.

To make a prediction we can feed a sample $n$ through the face and ECG identification models to retrieve the $\textbf{C}^{face}_{k,n}$ and $\textbf{C}^{ecg}_{k,n}$ for that particular sample. We then calculate \textit{fused confidence scores}, $\textbf{f}^{face}_{k,n}$ and $\textbf{f}^{ecg}_{k,n}$, using Equations \ref{eq:fusedScore1} and \ref{eq:fusedScore2}.

\begin{align} \label{eq:fusedScore1}
\textbf{f}^{face}_{k,n} = \textbf{C}^{face}_{k,n} \cdot \bigg(0.5 \cdot \textbf{J}_{1,M} - \textbf{D}_k\bigg) 
\\ \label{eq:fusedScore2}
\textbf{f}^{ecg}_{k,n} = \textbf{C}^{ecg}_{k,n} \cdot \bigg(0.5 \cdot \textbf{J}_{1,M} + \textbf{D}_k\bigg) 
\end{align}

$\textbf{f}_{k,n}$ exists for each model as $\textbf{f}^{face}_{k,n}$ and $\textbf{f}^{ecg}_{k,n}$ and follows the same subscript and superscript conventions as $\textbf{C}$. $0.5 \cdot \textbf{J}_{1,M}$ is an $M = 87$ length vector of containing $0.5$ in each element where $M = 87$ is the number of subjects. $(0.5 \cdot \textbf{J}_{1,M} \pm \textbf{D}_k)$ is an $M = 87$ length vector which contains values that are considered to be the weights for each subject. The better a model performs (relative to the other model), the larger the values in $(0.5 \cdot \textbf{J}_{1,M} \pm \textbf{D}_k)$ will be. $(0.5 \cdot \textbf{J}_{1,M} \pm \textbf{D}_k)$ creates a larger influence over the final decision for the model which performs better for any particular subject.

Equation \ref{eq:finalScore} describes the calculation of the \textit{final confidence score}, $\textbf{F}$. The predicted subject is for the fusion system is determined by finding $\argmax(\textbf{F})$. 

\begin{equation} \label{eq:finalScore}
\textbf{F}_{k,n} = \textbf{f}^{face}_{k,n} + \textbf{f}^{ecg}_{k,n}
\end{equation}

For pedagogical purposes, we will work through an example prediction of the fusion system using the trained model from the first fold. Figure \ref{fig:fusionAlgorithm} depicts the classification process of arbitrarily chosen test sample 3 from the fold 1 of the 10-fold cross-validation. The face and ECG data for test sample 3 is input to their respective classifiers to provide $\textbf{C}_f$ and $\textbf{C}_e$. The \textit{fused confidence scores}, \textbf{f}, are then calculated by finding the dot product of each $\textbf{C}$ vector with the result of $(0.5 \cdot \textbf{J} \pm \textbf{D})$. Finally, the fused confidence score vectors are added together to get the \textit{final confidence score} $\textbf{F}$. The predicted class, $\argmax_1(\textbf{F})$, is found to be subject 5.

For validation of our fusion system we once again use 10-fold cross-validation. This will allow us to directly compare the results of the face, ECG, and fused systems. We also test the fusion system performance under noisy data conditions by artificially generating poor quality samples by adding additive white gaussian noise to the face and ECG samples. The samples from every second and third subject for the face identification system are degraded with noise. The samples from every seventh subject for the ECG identification system are degraded. These subjects are chosen specifically to ensure the following conditions are present in the dataset:
\begin{itemize}
	\item There are subjects with good quality samples for both models (such as samples for subjects 1 and 5).
	\item There are subjects with good quality samples for the face model but poor quality samples for the ECG model (such as samples for subjects 7 and 35).
	\item There are subjects with good quality samples for the ECG model but poor quality samples for the face model (such as samples for subjects 2 and 3).
	\item There are subjects with poor quality samples for both models (such as samples for subjects 14 and 21).
\end{itemize}

\section{Results and Discussion} \label{sec:results}
\begin{table}[]
	\centering
	\caption{Summary of the 10-fold cross-validation results for the face, ECG, and proposed fusion identification systems.}
	\label{tab:accResults}
	\begin{tabular}{|c|c|c|c|}
		\hline
		\textbf{Fold} & \textbf{\begin{tabular}[c]{@{}c@{}}Face Test \\ Accuracy\\ (\%)\end{tabular}} & \textbf{\begin{tabular}[c]{@{}c@{}}ECG Test\\ Accuracy\\ (\%)\end{tabular}} & \textbf{\begin{tabular}[c]{@{}c@{}}Fusion Test\\ Accuracy\\ (\%)\end{tabular}} \\ \hline
1	&	98.966	&	96.897	&	100.000	\\	\hline
2	&	99.195	&	95.977	&	99.540	\\	\hline
3	&	98.506	&	98.161	&	99.540	\\	\hline
4	&	98.736	&	95.862	&	100.000	\\	\hline
5	&	98.966	&	94.023	&	99.655	\\	\hline
6	&	99.080	&	95.632	&	99.885	\\	\hline
7	&	98.736	&	96.437	&	99.770	\\	\hline
8	&	98.161	&	95.517	&	99.540	\\	\hline
9	&	98.966	&	96.897	&	99.885	\\	\hline
10	&	99.080	&	95.977	&	100.000	\\	\hline
		\textbf{avg $\pm$ std} & \textbf{98.839 $\pm$ 0.30} & \textbf{96.138 $\pm$ 1.02} & \textbf{99.782 $\pm$ 0.19} \\ \hline
	\end{tabular}
\end{table}

\begin{figure}[h!]
	\centering
	\includegraphics[width=\linewidth]{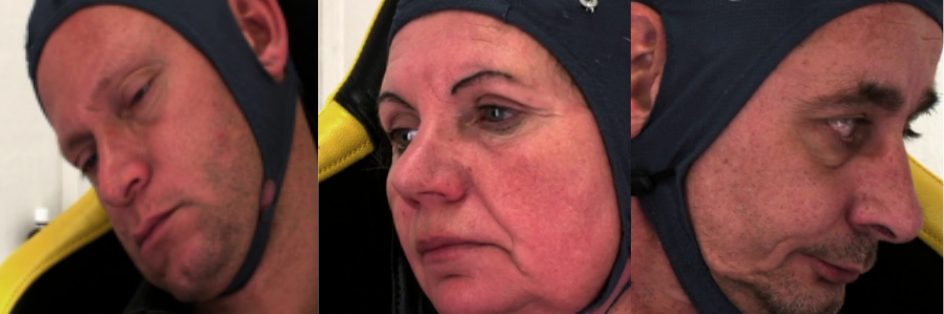} 
	\caption{Examples of poorly classified face samples.}
	\label{fig:Faces}
\end{figure}

\begin{figure}[h!] 
	\centerline{\includegraphics[width=\linewidth]{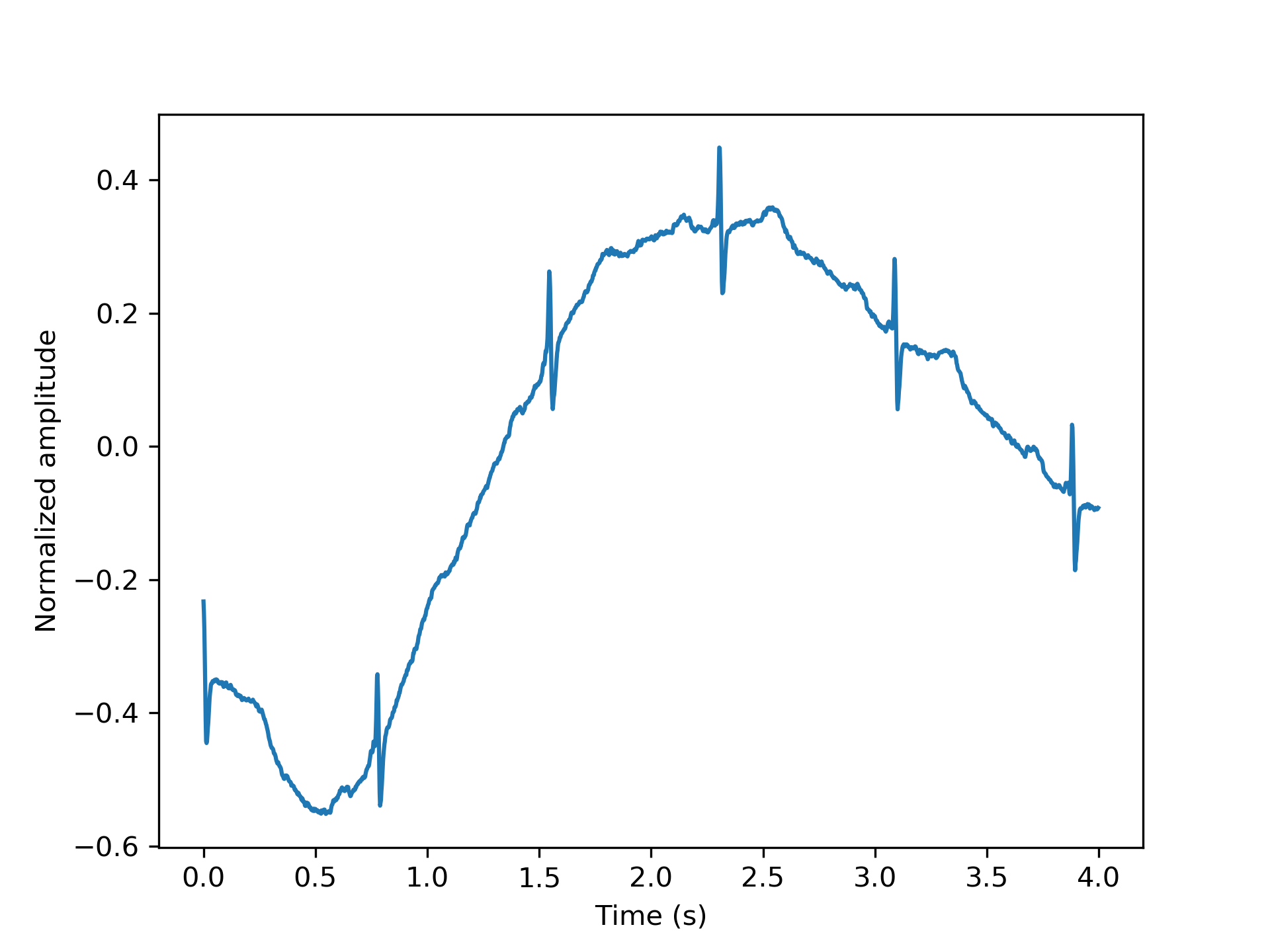}} 
	\caption{Example of a normalized and time-gated ECG signal that has been corrupted with a high power, low frequency noise.}
	\label{badECGSignal}
\end{figure}

Table \ref{tab:accResults} summarizes the results of the 10-fold cross-validation done for each of the identification models. The face identification model performed extremely well with an average test accuracy of 98.8\%. Errors in the face identification model were often caused by samples where the subject is turning their heads to extreme angles which obscures certain facial features and prevents the face detection algorithm from detecting and centering the face. Figure \ref{fig:Faces} shows a sample where the subject's head is tilted and turned at an extreme angle. The ECG identification system performed fairly well with an average test accuracy of 96.1\% accuracy. Errors from the ECG identification system often came from samples which contained a very high power, low frequency distortion. Figure \ref{badECGSignal} shows an example of a sample with this distortion. The reason for this distortion is unknown and the BioVid dataset does not comment on why these distortions occur. We hypothesize these distortions are likely a result of improper sensor connections to the subject.

The fusion identification system was able to improve the identification accuracy to 99.8\%. More specifically, of the 870 test samples for each fold, the fusion identification system averaged only about 1.85 errors for each fold. These results demonstrate the well known benefit of having multi-modal biometrics for identification. The novel combination of deep neural networks for face and ECG identification is able to perform extremely well on the BioVid dataset with well controlled data collection environments. The use of multiple sensors with mutually independent failure modes has been shown to significantly improve identification results.

Table \ref{tab:accNoisyResults} shows the results of the system under noisier data conditions. A comparison to the popular score-level fusion method, the weighted sum, as described in \cite{Jain2005}, is also included. The proposed fusion system was able to improve the individual accuracies of 66.6\% and 76.3\% to 93.8\%. Our proposed fusion method also exceeds the 91.5\% accuracy of the weighted sum method.

\begin{table}[]
	\centering
	\caption{Summary of the 10-fold cross-validation results for the face, ECG, proposed fusion, and weighted sum fusion systems.}
	\label{tab:accNoisyResults}
	\setlength\tabcolsep{1.5pt}
	\begin{tabular}{|c|c|c|c|c|}
		\hline
		\textbf{Fold} & \textbf{\begin{tabular}[c]{@{}c@{}}Face Test \\ Accuracy\\ (\%)\end{tabular}} & \textbf{\begin{tabular}[c]{@{}c@{}}ECG Test\\ Accuracy\\ (\%)\end{tabular}} & \textbf{\begin{tabular}[c]{@{}c@{}}Proposed \\ Fusion Test\\ Accuracy\\ (\%)\end{tabular}} &
		\textbf{\begin{tabular}[c]{@{}c@{}}Weighted \\ Sum Test\\ Accuracy\\ (\%)\end{tabular}}\\ \hline
1	&	69.195	&	81.149	&	94.368	&	92.644	\\	\hline
2	&	66.437	&	77.586	&	94.483	&	92.529	\\	\hline
3	&	66.552	&	78.391	&	94.023	&	91.839	\\	\hline
4	&	65.632	&	72.644	&	93.333	&	91.379	\\	\hline
5	&	66.552	&	72.184	&	91.609	&	88.736	\\	\hline
6	&	66.552	&	76.322	&	94.598	&	92.414	\\	\hline
7	&	67.931	&	76.782	&	94.368	&	91.379	\\	\hline
8	&	65.172	&	76.437	&	94.253	&	90.805	\\	\hline
9	&	65.862	&	77.241	&	93.103	&	91.724	\\	\hline
10	&	65.977	&	74.023	&	93.908	&	91.724	\\	\hline
		\textbf{avg $\pm$ std} & \textbf{66.586 $\pm$ 1.12} & \textbf{76.276 $\pm$ 2.57} & \textbf{93.805 $\pm$ 0.87} & \textbf{91.517 $\pm$ 1.08}\\ \hline
	\end{tabular}
\end{table}

\section{Conclusions and Future Works} \label{sec:conclusion}
The BioVid Heat Pain Database was used to create two separate identification models, one neural network model for face identification and one neural network model for ECG identification. A novel hybrid score- and rank-level fusion identification system was proposed and shown to improve identification results. Individually, using 10-fold cross-validation, the face identification model achieved an accuracy of 98.8\% and the ECG identification model achieved an accuracy of 96.1\% across 87 subjects. Using the same validation method, our fusion algorithm was able to achieve an accuracy of 99.8\%. To further test our proposed fusion algorithm, we artificially degrade the quality of the samples by adding additive white gaussian noise to the samples to reduce the identification rates of the face and ECG models. Our proposed fusion system was able to improve individual rates of 66.6\% and 76.3\% for the face and ECG models, respectively, to 93.8\%, a significant improvement over the individual systems. Longer term future work will include:
\begin{itemize}
	\item The proposed fusion algorithm is unsuitable for more than two modalities. Modifications to the proposed algorithm are needed to accommodate for more modalities.
	\item Applying the proposed methods to databases with noisier data to better analyze and understand the performance and sources of error for the proposed fusion system.
\end{itemize}

\section*{Acknowledgment}
This work was partially funded by the Natural Sciences and Engineering Research Council of Canada (NSERC Engage grant and CREATE grant "We-TRAC").

\bibliography{references}
\bibliographystyle{IEEEtran}

\end{document}